\documentclass{article}

\usepackage{arxiv}

\usepackage[utf8]{inputenc} 
\usepackage[T1]{fontenc}    
\usepackage{hyperref}       
\usepackage{url}            
\usepackage{booktabs}       
\usepackage{amsfonts}       
\usepackage{nicefrac}       
\usepackage{microtype}      
\usepackage{lipsum}		
\usepackage{graphicx}
\usepackage{natbib}
\usepackage{doi}

\usepackage{graphicx}
\usepackage{bm}
\usepackage{subcaption}
\usepackage{inconsolata}
\usepackage{amsmath}
\usepackage{cleveref}

\title{Revisiting Distance Metric Learning for Few-Shot Natural Language Classification}

\author{ \href{https://orcid.org/0000-0002-2241-9588}{\includegraphics[scale=0.06]{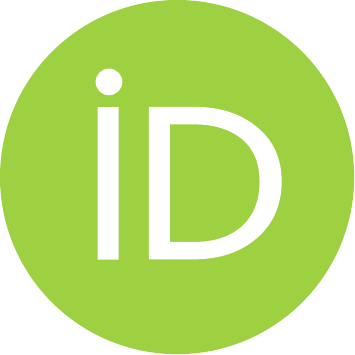}\hspace{1mm}Witold Sosnowski} \\
	Faculty of Mathematics and Information Science, \\
	Warsaw University of Technology, Poland \\
	\And
	\href{https://orcid.org/0000-0002-3407-7570}{\includegraphics[scale=0.06]{orcid.pdf}\hspace{1mm}Anna Wróblewska} \\
	Faculty of Mathematics and Information Science, \\
	Warsaw University of Technology, Poland \\
	\texttt{anna.wroblewska1@pw.edu.pl} \\
	\And
	\href{https://orcid.org/0000-0003-0617-7301}{\includegraphics[scale=0.06]{orcid.pdf}\hspace{1mm}Karolina Seweryn} \\
	Faculty of Mathematics and Information Science, \\
	Warsaw University of Technology, Poland \\
	NASK - National Research Institute, Poland \\
	\texttt{karolina.seweryn@pw.edu.pl} \\
	\And
	\href{https://orcid.org/0000-0002-9647-6761}{\includegraphics[scale=0.06]{orcid.pdf}\hspace{1mm}Piotr Gawrysiak} \\
	Faculty of Electronics and Information Technology \\
	Warsaw University of Technology, Poland \\
}

\renewcommand{\shorttitle}{Revisiting Distance Metric Learning for Few-Shot Natural Language Classification}

\hypersetup{
pdftitle={Revisiting Distance Metric Learning for Few-Shot Natural Language Classification},
pdfauthor={Witold Sosnowski, Anna Wróblewska, Karolina Seweryn, Piotr Gawrysiak},
}

\begin{document}
\maketitle

\begin{abstract}
	Distance Metric Learning (DML) has attracted much attention in image processing in recent years. This paper analyzes its impact on supervised fine-tuning language models for Natural Language Processing (NLP) classification tasks under few-shot learning settings.
    We investigated several DML loss functions in training RoBERTa language models on known SentEval Transfer Tasks datasets.  We also analyzed the possibility of using proxy-based DML losses during model inference.
    
    Our systematic experiments have shown that under few-shot learning settings, particularly proxy-based DML losses can positively affect the fine-tuning and inference of a supervised language model. Models tuned with a combination of CCE (categorical cross-entropy loss) and ProxyAnchor Loss have, on average, the best performance and outperform models with only CCE by about 3.27 percentage points -- up to 10.38 percentage points depending on the training dataset.

\end{abstract}


\section{Introduction}

Natural Language Processing (NLP) and Computer Vision (CV) are among the most important Machine Learning fields. The last breakthroughs resulted in models that share similar concepts, such as representation learning or transfer learning~\cite{le2020contrastive, zhuang2020comprehensive}. It is sufficient to mention ResNet~\cite{he2016deep} or DenseNet~\cite{huang2017densely} in the CV and BERT~\cite{devlin2018bert} or RoBERTa~\cite{roberta} from the NLP field. These models are preliminarily pre-trained on the massive amount of general data and can be fine-tuned to suit the downstream task such as classification or entailment. In the field of CV, there is a very lively discussion about using Metric Learning methods for fine-tuning pre-trained models. Metric Learning tries to shape the embedding space so that similar data are close to each other while dissimilar are far from each other~\cite{kaya2019deep}. 

In NLP, the cross-entropy loss has been widely used for supervised fine-tuning language models. However, it turned out that it is not always the most optimal objective function since it is shown to have relatively high variations across multiple runs with different seeds, even though only a few training components can relate directly to the seeds~\cite{zhang2020revisiting}. It is even more unstable in fine-tuning models where the number of observations is relatively limited~\cite{bansal2019learning}. 

The DML methods address some of the CE problems. Instead of learning to distinguish observations from different classes, the DML losses are meant to minimize the distances between representations of observations from the same class and maximize distance if the classes are different. In this way, it learns not separable features of the representations as in the case of CE loss but how the general representation of the class looks~\cite{wen2016discriminative}. The difference between the CE loss and DML is especially visible when the number of observations is small. 

Therefore, this research aims to create a benchmark that shows how different DML losses can influence supervised fine-tuning language models in the few-shot learning settings in NLP. Previous work of Gunel and team~\cite{gunel2020supervised} first examined how to combine Cross-Entropy and Supervised Contrast Learning losses for the RoBERTa-large fine-tuning. Still, it has certain limitations, e.g. it only considered one of many possible DML losses and did not use the entire language model output, only the vector associated with the token [CLS]. We extend these experiments by introducing more advanced proxy-based DML losses and applying them to the entire RoBERTa output, which has been comprehensively analyzed based on the 40-fold cross-validation. We have also developed an inference method incorporating proxy-based DML losses into the inference/prediction process.

This paper's contributions are the following:
\begin{enumerate}
    \item We have described the language model fine-tuning process with DML loss functions and added modifications that improve the method for applying proxy-based DML losses.
    \item We have elaborated an inference method that relies on proxies derived from proxy-based DML methods and can further improve performance on downstream tasks.
    \item We have conducted systematic and thorough experiments based on 40-fold cross-validation over several datasets to investigate how different DML losses perform in the few-shot learning settings for the supervised fine-tuning of the RoBERTa-base and RoBERTa-large language models.
\end{enumerate}

The following Section~\ref{sec:background} is a brief overview of the DML methods. Section~\ref{sec:method} outlines our approach -- a generalization of DML in training, modifications in inference and experimental procedure. The next Section~\ref{sec:experiments} provides a performance analysis of the models and investigates their behaviour. Finally, a summary of the experiments is described in the last Section~\ref{sec:conclusion}.

\section{Methodological Background}\label{sec:background}

As already stated, Distance Metric Learning involves transforming the embedding space so that representations of observations from the same class are close together, while those from different classes are far apart~\cite{qian2019softtriple}. Most often, the DML is used when the task is related to the information retrieval, such as retrieving data that are most similar to a query~\cite{musgrave2020metric}, during self-supervised visual representation training~\cite{he2020momentum}, k-nearest neighbours classification~\cite{weinberger2009distance} or clustering\cite{xing2002distance}.

Throughout this paper, we use the following terms. Observations, which we refer to as $x_{i}$ are represented by embeddings $z_{i}$, where $i \in I \equiv\{1, \ldots, N\}$ indicates the index of the anchor observation $x_{i}$. $x^{a}$ denotes an anchor, and $z^{a}$ stands for its vector representation. $x^{p}$ belongs to the same class as the anchor (positive observation) and $x^{n}$ - the negative observation, is from different than the anchor class.

\subsection{Triplet Loss}

One of the most popular DML methods is Triplet Loss, which is the extension of the Contrastive Loss \cite{schroff2015facenet}. It calculates loss from representations triplets of the sample ${x^{a}}, {x^{p}}, {x^{n}}$ 
The main idea is to minimize the distance between the representations of the anchor and positive observations while maximizing between an anchor and a negative one. The Triplet Loss is given in Equation~\ref{eq:triplet-loss}.

\begin{equation}\label{eq:triplet-loss}
\ell_{Triplet}=\sum_{i=1}^{N}\left[\left\|z_{i}^{a}-z_{i}^{p}\right\|_{2}^{2}-\left\|z_{i}^{a}-z_{i}^{n}\right\|_{2}^{2}+m\right]_{+},
\end{equation}

\noindent where 
$m$ stands for a margin imposed between positive and negative examples. 


\subsection{N-Pairs Loss}
Another important loss DML is the \textit{N-Pairs loss}~\cite{sohn2016improved}. It represents a minor evolution of the Triplet Loss in that it computes the loss not only from a triplet containing the anchor, single positive and single negative sample representations but pairs the anchor with the single positive and all negative observation representations. The \textit{N-Pair Loss} is given in the Equation~\ref{eq:n-pair-loss}.

\begin{equation}\label{eq:n-pair-loss}
\resizebox{0.50\textwidth}{!}{
$\ell_{NPairs}=\frac{-1}{N} \sum\limits_{i=1}^{N} \log \frac{\exp \left(z_{i}^{a} \cdot z_{i}^{p}\right)}{\exp \left(z_{i}^{a} \cdot z_{i}^{p}\right)+\sum\limits_{j \neq i} \exp \left(z_{i}^{a} \cdot z_{j}^{n}\right)}
$}
\end{equation}


\subsection{Supervised Contrastive Learning Loss}
Supervised Contrastive Learning (\textit{SupCon}) is one of the modern non-proxy DML approaches; it most often outperforms the other non-proxy DML methods such as Triplet Loss or Contrastive Loss. The \textit{SupCon} introduces temperature regularization~\cite{pereyra2017regularizing} and batch processing which means that the loss is not calculated just for a single triplet but is the average of all possible triplets from a given batch.
The \textit{SupCon} loss extends the self-supervised batch DML approach to the fully-supervised setting, which provides the ability to use the label information~\cite{khosla2020supervised}. Instead of contrasting one positive example for an anchor with all other observations from the batch, \textit{SupCon} contrasts all examples from the same class (as positives) with all other observations from the batch as negatives. The most critical issue with this approach is that as the number of observations in the batch grows, the number of triplets grows cubically.

The \textit{SupCon} loss has been previously applied in the supervised fine-tuning of the RoBERTa-large language model~\cite{gunel2020supervised} and its formula is as follows:
\begin{equation} \label{eq:supcon-loss}
\resizebox{0.5\textwidth}{!}{
$\ell_{SupCon}=\sum\limits_{i \in I} \frac{-1}{|P(i)|} \sum\limits_{p \in P(i)} \log \frac{\exp \left(\boldsymbol{z}_{i} \cdot \boldsymbol{z}_{p} / \tau\right)}{\sum\limits_{k \in K(i)} \exp \left(\boldsymbol{z}_{i} \cdot \boldsymbol{z}_{k} / \tau\right)},$
}
\end{equation}

\noindent where $i \in I \equiv\{1, \ldots, N\}$ denotes the index of an anchor observation $x_i$, 
$P(i)$ is the set of observations from the same class that $x_i$ belongs to, $K(i) \equiv I \backslash\{i\}$, 
and $\tau \in \mathcal{R}^{+}$ is a scalar temperature parameter.

\subsection{ProxyNCA Loss}
\textit{ProxyNCA Loss} is the first loss we have used that is proxy-based~\cite{movshovitz2017no}. Proxies are artificial embeddings that can be learned in the training process and are designed to represent data. In supervised settings, there are as many proxies as the number of classes, so each class is well approximated by one artificial proxy. The loss is calculated based on possible triples formed from observations representing the anchor class, a positive proxy (a proxy from the anchor class) and a negative proxy (a proxy from a class different from the anchor). The advantage of the proxy-based method over previous approaches is that the number of all possible triples grows linearly to the cardinality of the dataset, whereas it grows cubically in the case of non-proxy-based loss. In addition, the proxies are more robust to outliers and thus have better performance in terms of speed and convergence. The \textit{ProxyNCA Loss} is given by the Equation~\ref{eq:proxy-nca-loss}:

\begin{equation}\label{eq:proxy-nca-loss}
\ell_{ProxyNCA}=-\frac{1}{N} \sum_{i=1}^{N} \log \left(\frac{e^{-d\left(z_{i}^{a}, p_{i}\right)}}{\sum\limits_{j=1, j \neq i}^{M} e^{-d\left(z_{i}^{a}, p_{j}\right)}}\right)
\end{equation}

\noindent where  
$j \in J \equiv\{1, \ldots, M\}$ denotes the index of proxies, $p_{j}$ represents the proxy of the class $j$ and $d$ denotes Euclidean distance.

\subsection{SoftTriple Loss}

\textit{SoftTriple} is based on a similar idea as \textit{ProxyNCA Loss} but provides the ability to introduce more than one proxy per class, so it can better reflect the distribution of features across classes~\cite{qian2019softtriple}. 

The following formulas define the \textit{SoftTriple} loss:
\begin{equation}\label{eq:softtriple-loss}
\resizebox{0.45\textwidth}{!}{
$\ell_{SoftTriple}
=-\frac{1}{N} \sum\limits_{i \in I}{\log \frac{\exp \left(\lambda\left(\mathcal{S}_{i, y_{i}}^{\prime}-\delta\right)\right)}{\exp \left(\lambda\left(\mathcal{S}_{i, y_{i}}^{\prime}-\delta\right)\right)+\sum\limits_{j \neq y_{i}} \exp \left(\lambda \mathcal{S}_{i, j}^{\prime}\right)}}$
}
\end{equation}

\begin{equation} \label{eq:softtriple-loss2}
\mathcal{S}_{i, c}^{\prime}=\sum_{k \in K} \frac{\exp \left(\frac{1}{\gamma} {{z}_{i}}^{\top} {w}_{c}^{k}\right)}{\sum\limits_{k \in K} \exp \left(\frac{1}{\gamma} {{z}_{i}}^{\top} {w}_{c}^{k}\right)} {{z}_{i}}^{\top} {w}_{c}^{k},
\end{equation}

\noindent where 
$y_i$ is 
label of $x_{i}$, $c \in C$ denotes the class index, $C$ is a set of class indices, 
$k \in K$ denotes the proxy index of class $c$, $K$ is a set of proxies per class, 
$\delta$ denotes a minimum interclass margins, $\lambda$ scaling factor that reduces the outliers' impact, $\gamma$ scaling factor for the entropy regularizer,  
$E(\cdot) \in \mathbf{R}^{d}$ denotes an encoder and $\mathbf{w}_{c}^{k}$ are weights that represent embeddings of the class $c$.

\subsection{AnchorProxy Loss}

\textit{ProxyAnchor Loss}, unlike the \textit{ProxyNCA loss}, represents the anchor as an approximation, while positive observations and negative classes are represented as embeddings~\cite{kim2020proxy}.
Equation~\ref{eq:proxy-anchor-loss} defines the \textit{AnchorProxy Loss}.

\begin{equation}\label{eq:proxy-anchor-loss}
\resizebox{0.45\textwidth}{!}{
$\begin{aligned} \ell_{\text {AnchorProxy }} &=\frac{1}{\left|P_{+}\right|} \sum_{p \in P_{+}} \log \left(1+\sum_{z \in Z_{p}^{+}} e^{-\alpha(s(z, p)-\delta)}\right) \\ &+\frac{1}{|P|} \sum_{p \in P} \log \left(1+\sum_{z \in Z_{p}^{-}} e^{\alpha(s(z, p)+\delta)}\right) \end{aligned}$
}
\end{equation}

\noindent where $P$ denotes the set of all proxies, $P_{+}$ denotes the set of all proxies from the batch (proxies associated with classes represented by batch observations), $Z_{p}^{+}$ denotes batch representations from the same class as the proxy $p$, $Z_{p}^{-}$ represents embeddings from different than $p$ classes, $\alpha > 0$ denotes a scaling factor, $\delta > 0$ represents a margin and $s(\cdot,\cdot)$ is the cosine similarity between two vectors.

\section{Our Approach}
\label{sec:method}
We study the influence of the DML loss function on fine-tuning a pre-trained language model by extending the categorical cross-entropy (CCE) loss with different DML loss functions. We also conduct experiments on the effect of reusing trained proxies from proxy-based DML methods in the inference process.

\subsection{DML for Loss Function}
In our experiments with fine-tuning models, we utilize a loss function that combines both the \textit{categorical cross-entropy} loss and the \textit{DML}, as given in Equation~\ref{eq:novel-loss}.

\begin{equation}
\mathcal{L}=(\beta) \mathcal{\ell}_{CCE}+(1 - \beta) \mathcal{\ell}_{DML},
\label{eq:novel-loss}
\end{equation}

\begin{equation}
\mathcal\ell_{CCE}=-\frac{1}{N} \sum_{i=1}^{N} \sum_{c=1}^{C} y_{i, c} \cdot \log \hat{y}_{i, c}
\end{equation}

\noindent where
$N$ denotes the observations number,
$C$ is the class number,
$y_{i c}$ represents the label of the $i$th observation from the $cth$ class,
$p_{i c}$ represents the model prediction for the $i$th observation from the $c$th class,
$\beta$ denotes the scaling factor that tunes the influence of both parts of the loss,
$\ell_{{DML }}$ denotes DML losses described above such as $\ell_{Triplet}$, $\ell_{NPairs}$, $\ell_{SupCon}$, $\ell_{ProxyNCA}$, $\ell_{SofTriple}$ or $\ell_{AnchorProxy}$.
$\ell_{{CCE }}$ stands for the \textit{categorical cross-entropy} loss.

\begin{figure}[!ht]
    \centering
    \includegraphics[scale=0.5]{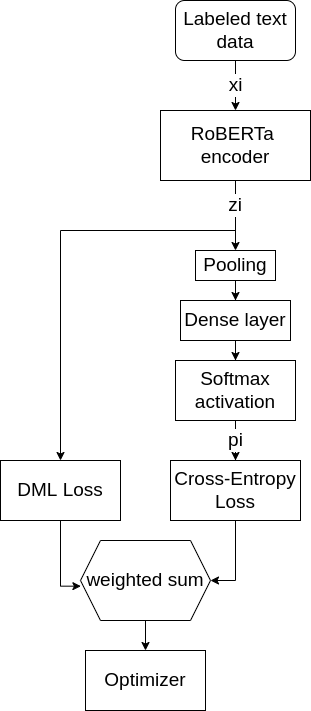}
    \caption{RoBERTa-based model fine-tuning architecture with the loss function as the weighted sum of DML Loss and Cross-Entropy Loss.}\label{fig:training-new-loss-function-overview}
\end{figure}

The first case is not possible for non proxy-based DML functions (\textit{Triplet Loss}, \textit{N-Pairs Loss} and \textit{SupCon Loss}) due to the high resource intensity. Figure~\ref{fig:training-new-loss-function-overview} sketches the whole training procedure.

\subsection{Proxy-based DML for Inference}
\begin{figure}[!ht]
    \centering
    \includegraphics[scale=0.5]{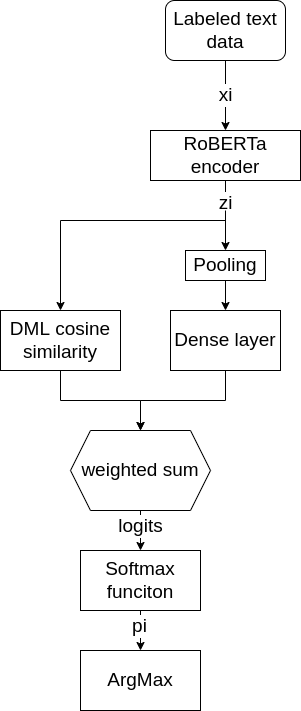}
    \caption{RoBERTa-based model inference architecture. The logits are designed as the weighted sum of the dense layer logits and cosine similarity between the observation representation $z_{i}$ and the DML proxies.}\label{fig:inference-new-logits-function-overview}
\end{figure}

In inference experiments, we explore the effect of reusing trained proxies from the proxy-based DML methods on the inference process. We extend the original vector of logits\footnote{The logits are defined as the vector of non-normalized predictions that a classification layer generates} derived from the dense classifier layer with additional logits computed as the cosine distance between the observation representation outputted by the encoder to the proxy from the DML loss -- see Equation~\ref{eq:novel-logits}.

\begin{equation}
{p_{i_c}}=\beta {\sigma(dense(z_{i}))}+(1 - \beta){s(z_{i}, pr_{c})},
\label{eq:novel-logits}
\end{equation}

\noindent where $i$ denotes the index of the observation $x_{i}$, $p_{i_c}$ denotes the probability of the observation $x_{i}$ belonging to the class $c$, $pr_{c}$ denotes the proxy belonging to the class $c$, $dense()$ denotes the dense layer that maps the l-dimensional embedding $z_{i}$ to the $K$ dimensional space, where $K$ is the number of classes, $\sigma()$ is the softmax function, $\beta$ is the scaling factor that tunes the influence of both parts of the inference function and $s()$ denotes the similarity function between two vectors - in our case it is the cosine similarity. Figure~\ref{fig:inference-new-logits-function-overview} sketches the modified inference process.


\subsection{Experimental Procedure}
As the language model we use the current state-of-the-art RoBERTa~\cite{liu2019roberta} encoder provided by the \textit{huggingface} library as the pre-trained model in the large and base form: respectively \textit{roberta-large} and \textit{roberta-base}. The standard procedure for supervised fine-tuning of the RoBERTa-based language models starts by tokenizing the input text $x$ to the array of numbers with the special tokens such as \textit{[CLS]} the first token, \textit{[EOS]} the end token and \textit{[SEP]} as the separating sentences token. The tokenized text is then fed to the RoBERTa encoder, which outputs the embedding array $z$. It is then passed to the dense layer $dense()$, which returns the logits array $l$, which cardinality equals the number of classes. The output is passed to the softmax function $\sigma()$, and then to the \textit{categorical cross-entropy} function that calculates the loss.

The proxy-based DML methods are fed with the entire output of RoBERTa models, while non-proxy-based methods are much more resource intensive as they need to compare all observations from the training sets and are therefore fed with embeddings associated with the token \textit{[CLS]}, as proposed in the paper~\cite{gunel2020supervised}. 

We test our models in the 40-fold cross-validation settings. It implies that each result is an average F1 score of 40 runs. In line with the conclusions of the paper~\cite{gunel2020supervised}, we restricted our study to the few-shot learning settings limiting the datasets to 20, 100 and 1,000 observations. For each dataset, we generated 40 folds with the same seed for different test models. Each of the 40 folds consisted of training and test sets, from which we sampled the training set with the same seed so that it was limited to 20, 100, or 1,000 observations in different experiments. It ensured that each test model was trained and tested on the same data. The best hyperparameters were chosen based on the model with the best average F1 score. 
Although the 40-fold cross-validation is very time-consuming, we decided to apply it to tackle the problem of high variance, which is common when the amount of training data is limited~\cite{dvornik2019diversity}. 

For each dataset, we compare our results with the baseline which results were separately obtained based on the hyperparameter search with the same batch size $=64$, learning rates $\in\{1e-5, 2e-5, 3e-5\}$, epochs number $\in\{8,16,64,128\}$, linear warmup for the first 6\% of steps and weight decay coefficient $=0.01$.
The final best hyperparameters for the baselines are the following: the learning rate of $1e-5$ for each dataset, and $8$ epochs for 1,000 elements datasets, $64$ for 100-element datasets and $128$ for 20-element datasets.

Each DML loss has its own hyperparameter search. The search space of hyperparameter $\beta$ was the same for all methods: $\beta \in\{0.1, 0.3, 0.5, 0.7, 0.9\}$. Apart from that, there are parameters depending on the method.
For the \textit{Triplet Loss} the grid search included additional hyperparameter 
$m \in\{1, 3 ,5, 7, 9\}$. 
The model with the \textit{SupCon Loss} was optimised based on 
$\tau \in\{0.1, 0.3, 0.5, 0.7, 0.9\}$
. In the case of \textit{SoftTriple} loss, we searched the following parameters: $k \in\{5, 25, 1,000, 2000\}$, $\gamma \in\{0.01, 0.03, 0.05, 0.07, 0.1\}$, $\lambda \in\{1,3,3.3,4,6,\\8,10\}$, and $\delta \in\{0.1, 0.3, 0.5, 0.7, 0.9, 1\}$. 
The \textit{ProxyNCA Loss} was optimised according to the grid search that included 
$softmax scale \in\{0.4, 0.6, 0.8, 1, 1.2, 1.4, 1.6, 1.8, 2, 3, 5\}$.
The \textit{ProxyAnchor Loss} was optimised based on the following additional hyperparameters: 
$\alpha \in\{16, 32, 64, 128\}$ and $\delta \in\{0, 0.1, 0.3, 0.5, 0.7, 0.9\}$.

We conducted experiments on the SentEval Transfer Task datasets from the classification and textual entailment tasks~\cite{conneau2018senteval}, as described in  
Table~\ref{tab:datasets_description}. 

\begin{table}[!ht]
\centering
\caption{SentEval Transfer Task datasets used in our experiments.}
\label{tab:datasets_description}
\resizebox{1\columnwidth}{!}{%
\begin{tabular}{llll}
\hline
\textbf{Dataset} &\textbf{\#Sentences} & \textbf{\#Classes} & \textbf{Task}\\
\hline
SST2 & 67k & 2 &  Sentiment (movie reviews)\cite{socher2013recursive}\\
MR & 11k & 2 &  Sentiment (movie reviews) \cite{pang2005seeing} \\
MPQA & 11k & 2 &  Opinion polarity \cite{wiebe2005annotating}\\
SUBJ & 10k & 2 &  Subjectivity status \cite{pang2004sentimental}\\
TREC & 5k & 6 &  Question-type classification \cite{pang2005seeing}\\
CR & 4k & 2 &  Sentiment (product review) \cite{hu2004mining}\\
\end{tabular}
}
\end{table}



\begin{table*}[h!]
\centering
\caption{F1 score of RoBERTa-base (RB) vs RoBERTa-base with DML losses trained on the 20-element datasets.}
\resizebox{\columnwidth}{!}{%
\begin{tabular}{llllllll}
\hline
\textbf{Loss} &                      \textbf{SST2}                  & \textbf{MR}                  & \textbf{MPQA}         & \textbf{SUBJ}         & \textbf{TREC}         & \textbf{CR}          &  \textbf{Avg}\\
\hline
CCE&60.04$\pm$8.49&62.01$\pm$8.61&65.41$\pm$5.38&87.05$\pm$4.38&39.43$\pm$10.61&67.99$\pm$9.39&63.66\\
CCE + Triplet &63.93$\pm$7.49*&64.74$\pm$6.99&64.57$\pm$5.45&\textbf{88.66}$\pm$3.09&40.93$\pm$9.41&\textbf{72.93}$\pm$8.95*&65.96\\
CCE + NPairs &66.50$\pm$8.56*&65.27$\pm$7.31&64.78$\pm$4.79&87.47$\pm$3.08&42.55$\pm$10.11&67.78$\pm$7.98&65.72\\
CCE + SupCon &69.15$\pm$7.32*&65.38$\pm$7.96&\textbf{68.99}$\pm$8.06*&87.68$\pm$2.80&42.40$\pm$8.87&70.16$\pm$10.51&67.29\\
CCE + SoftTriple &64.20$\pm$8.53*&\textbf{68.83}$\pm$7.08*&68.44$\pm$5.52*&88.16$\pm$5.30&43.27$\pm$9.61&71.48$\pm$8.98&67.39\\
CCE + ProxyNCA
&65.30$\pm$7.13*&66.00$\pm$6.14*&64.81$\pm$4.26&86.86$\pm$5.00&\textbf{44.70}$\pm$9.57*&66.40$\pm$6.58&65.68\\
CCE + ProxyAnchor &67.50$\pm$4.87*&65.30$\pm$6.56&67.26$\pm$5.31&87.06$\pm$2.03&41.85$\pm$8.82&68.55$\pm$7.53&66.09\\
CCE + SoftTriple + inf &64.25$\pm$8.51*&67.33$\pm$7.82*&68.77$\pm$5.45*&88.37$\pm$4.81&43.40$\pm$9.58&71.04$\pm$8.68&67.19\\
CCE + ProxyNCA + inf 
&68.04$\pm$4.81*&65.17$\pm$6.69&67.61$\pm$5.27&86.14$\pm$5.38&41.85$\pm$8.82&71.31$\pm$6.83&66.69\\
CCE + ProxyAnchor + inf &\textbf{69.08}$\pm$6.20*&65.71$\pm$6.65*&68.36$\pm$6.61*&87.85$\pm$2.58&41.98$\pm$8.72&72.45$\pm$7.36*&\textbf{67.57}\\
\end{tabular}
}
\label{tab:rb-20-examples-dataset}
\end{table*}

\begin{table*}[h!]
\centering
\caption{F1 score of RoBERTa-base (RB) vs RoBERTa-base with DML losses trained on the 100-element datasets.}
\resizebox{\columnwidth}{!}{%
\begin{tabular}{llllllll}
\hline
\textbf{Loss} &                      \textbf{SST2}                  & \textbf{MR}                  & \textbf{MPQA}         & \textbf{SUBJ}         & \textbf{TREC}         & \textbf{CR}          &  \textbf{Avg}\\
\hline
CCE&83.08$\pm$3.52&80.57$\pm$3.53&81.64$\pm$5.10&92.02$\pm$1.95&75.99$\pm$4.20&\textbf{88.65}$\pm$3.37&83.66\\

CCE + Triplet &84.28$\pm$2.65&\textbf{82.24}$\pm$2.00*&74.81$\pm$5.32*&92.24$\pm$1.53&74.91$\pm$4.35&88.64$\pm$3.75&82.85\\
CCE + NPairs &83.72$\pm$4.25&81.48$\pm$1.82&82.00$\pm$4.04&92.20$\pm$1.29&75.60$\pm$3.97&85.85$\pm$4.20*&83.48\\
CCE + SupCon &85.29$\pm$1.71*&81.87$\pm$1.96*&\textbf{85.52}$\pm$2.51*&92.38$\pm$1.38&76.67$\pm$3.98&87.83$\pm$3.56&\textbf{84.92}\\
CCE + SoftTriple &84.42$\pm$2.55&81.31$\pm$3.15&81.43$\pm$6.16&{92.59}$\pm$1.98&{78.01}$\pm$4.37*&{88.48}$\pm$3.66&84.37\\
CCE + ProxyNCA &84.32$\pm$2.52&81.27$\pm$3.08&83.07$\pm$4.07&92.11$\pm$1.68&76.88$\pm$4.64&88.07$\pm$3.63&84.29\\
CCE + ProxyAnchor &84.69$\pm$1.84*&80.27$\pm$2.51&83.30$\pm$4.04&92.11$\pm$1.87&76.21$\pm$4.37&87.72$\pm$3.72&84.05\\
CCE + SoftTriple + inf &84.42$\pm$2.55&81.29$\pm$3.15&81.44$\pm$6.12&92.69$\pm$1.92&\textbf{78.86}$\pm$4.26*&88.41$\pm$3.56&84.52\\
CCE + ProxyNCA + inf &84.35$\pm$2.39&81.39$\pm$3.10&83.06$\pm$4.08&92.12$\pm$1.68&76.94$\pm$4.63&87.98$\pm$3.61&84.31\\
CCE + ProxyAnchor + inf &\textbf{85.58}$\pm$2.41*&81.89$\pm$2.05*&83.68$\pm$2.61*&\textbf{92.70}$\pm$0.88*&76.63$\pm$4.27&88.36$\pm$2.24&84.81\\
\end{tabular}
}
\label{tab:rb-100-examples-dataset}
\end{table*}

\begin{table*}[h!]
\centering
\caption{F1 score of RoBERTa-base (RB) vs RoBERTa-base with DML losses trained on the 1,000-element datasets.}
\resizebox{\columnwidth}{!}{%
\begin{tabular}{llllllll}
\hline
\textbf{Loss} &                      \textbf{SST2}                  & \textbf{MR}                  & \textbf{MPQA}         & \textbf{SUBJ}         & \textbf{TREC}         & \textbf{CR}          &  \textbf{Avg}\\
\hline
CCE&88.73$\pm$0.97&86.27$\pm$2.22&88.65$\pm$1.89&94.53$\pm$1.42&86.31$\pm$3.31&91.87$\pm$3.01&89.39\\
CCE + Triplet &88.46$\pm$0.68&86.91$\pm$1.62&87.38*$\pm$1.48&94.69$\pm$0.90&84.21$\pm$2.56*&92.02$\pm$2.01&88.95\\
CCE + NPairs &88.80$\pm$0.63&86.30$\pm$1.28&88.63$\pm$1.49&94.57$\pm$1.04&83.70$\pm$2.10*&92.11$\pm$2.27&89.02\\
CCE + SupCon &89.07$\pm$0.70&86.76$\pm$1.63&89.37$\pm$1.41&94.74$\pm$1.02&84.98$\pm$2.29*&\textbf{92.55}$\pm$2.21&89.58\\
CCE + SoftTriple &88.96$\pm$0.85&86.76$\pm$2.05&89.33$\pm$1.90&94.63$\pm$1.29&{87.19}$\pm$3.16&92.46$\pm$3.20&{89.89}\\
CCE + ProxyNCA &89.30$\pm$0.99*&86.70$\pm$1.88&88.83$\pm$2.05&94.69$\pm$1.47&83.72$\pm$3.46*&92.38$\pm$2.87&89.27\\
CCE + ProxyAnchorLoss &89.31$\pm$0.85*&86.56$\pm$2.20&89.26$\pm$1.87&94.40$\pm$1.56&81.10$\pm$3.77*&92.48$\pm$2.57&88.85\\
CCE + SoftTriple + inf &88.96$\pm$0.85&\textbf{86.77}$\pm$2.05&\textbf{89.42}$\pm$2.00&94.51$\pm$1.31&\textbf{87.42}$\pm$3.01 & 92.39$\pm$2.95&\textbf{89.91}\\
CCE + ProxyNCA + inf &89.31*$\pm$0.98&86.65$\pm$2.25&88.84$\pm$1.83&94.68$\pm$1.48&83.70$\pm$3.46*&92.30$\pm$2.85&89.25\\
CCE + ProxyAnchor + inf &\textbf{89.93}$\pm$0.93*&\textbf{86.77}$\pm$1.97&89.27$\pm$1.87&\textbf{95.04}$\pm$1.37&85.09$\pm$3.78&{92.48}$\pm$2.57&89.76\\
\end{tabular}
}
\label{tab:rb-1000-examples-dataset}
\end{table*}

\begin{table*}[h!]
\centering
\caption{F1 score of RoBERTa-large (RL) vs RoBERTa-large with DML losses trained on the 20-element datasets.}
\resizebox{\columnwidth}{!}{%
\begin{tabular}{llllllll}
\hline
\textbf{Loss} &                      \textbf{SST2}                  & \textbf{MR}                  & \textbf{MPQA}         & \textbf{SUBJ}         & \textbf{TREC}         & \textbf{CR}          &  \textbf{Avg}\\
\hline
CCE &   53.71$\pm$8.74          & 56.55$\pm$8.67 & 65.66$\pm$5.01 & 85.54$\pm$6.38 & 41.48$\pm$9.46 & 65.03$\pm$8.26 &  61.33\\
CCE + Triplet  &    60.77$\pm$10.06 *&  65.73$\pm$9.17* &68.01$\pm$7.29 & 88.98$\pm$2.83 *& 41.24$\pm$11.44&65.48$\pm$7.57 & 65.03\\
CCE + NPairs  &    57.69$\pm$9.77 &  65.19$\pm$10.57* & 64.89$\pm$4.80 & 87.33$\pm$5.16 &42.16$\pm$9.86&62.46$\pm$7.92 &63.29\\
CCE + SupCon  &     60.04$\pm$8.98 *         & 65.74$\pm$9.69* & 64.92$\pm$4.91 & 87.66$\pm$4.76 &    42.81$\pm$10.55        & 66.59$\pm$6.64 & 	64.63\\
CCE + SoftTriple  & {62.35}$\pm$7.44 *& {70.03}$\pm$8.16* & {67.96}$\pm$4.72* & {89.48}$\pm$4.62* & {47.21}$\pm$9.44 *& {68.55}$\pm$6.91 * &{67.60} \\
CCE + ProxyNCA  &  65.26$\pm$9.02* & 71.11$\pm$7.89*  & 70.09$\pm$6.41* & 90.04$\pm$3.61*&43.21$\pm$10.27 &69.16$\pm$8.57*&68.15\\
CCE + ProxyAnchorLoss  &   65.90$\pm$6.93* &70.53$\pm$9.87* &71.70$\pm$6.16* &89.72$\pm$2.86*&43.87$\pm$10.39&72.61$\pm$8.34*&69.06\\
CCE + SoftTriple + inf  &  64.81$\pm$10.76*  &68.95$\pm$12.00* & 68.98$\pm$6.04* &89.64$\pm$3.68* & 45.84$\pm$10.25&69.45$\pm$8.88* &67.95\\
CCE + ProxyNCA + inf  &   64.34$\pm$10.76*  &70.11$\pm$10.92* & 67.48$\pm$6.24 & 88.16$\pm$5.49 & 45.20$\pm$10.28&66.40$\pm$8.64&66.95\\
CCE + ProxyAnchorLoss + inf &   \textbf{68.50}$\pm$6.93*  &\textbf{72.25}$\pm$7.85* & \textbf{73.40}$\pm$6.10 *& \textbf{91.18}$\pm$1.34 *& \textbf{51.45}$\pm$7.00*&\textbf{73.45}$\pm$8.26*&\textbf{71.71}\\
\end{tabular}
}
\label{tab:rl-20-examples-dataset}
\end{table*}

\begin{table*}[h!]
\centering
\caption{F1 score of RoBERTa-large (RL) vs RoBERTa-large with DML losses trained on the 100-element datasets.}
\resizebox{\columnwidth}{!}{%
\begin{tabular}{llllllll}
\hline
\textbf{Loss} &                      \textbf{SST2}                  & \textbf{MR}                  & \textbf{MPQA}         & \textbf{SUBJ}         & \textbf{TREC}         & \textbf{CR}           &  \textbf{Avg}\\
\hline
CCE               & 85.87$\pm$5.50          & 82.57$\pm$5.94       & 82.50$\pm$5.65  & 93.91$\pm$1.47 & 82.72$\pm$4.73 & 89.43$\pm$3.65     & 86.16\\
CCE + Triplet  &87.73$\pm$2.12   &84.57$\pm$3.50 &81.45$\pm$7.73&93.68$\pm$1.65&78.64$\pm$6.48*&89.95$\pm$3.70&86.00\\
CCE + NPairs   &87.79$\pm$2.29*  &84.99$\pm$3.00*&78.44$\pm$5.93* &93.93$\pm$1.42&80.54$\pm$4.42*&89.10$\pm$5.26&85.79\\
CCE + SupCon      & {88.47}$\pm$1.38 * &85.24$\pm$3.30*      & 81.18$\pm$6.16   & 94.39$\pm$1.43  & 81.82$\pm$4.51 & 90.73$\pm$3.02&86.97 \\
CCE + SoftTriple  & 88.12$\pm$1.83*       & {85.49}$\pm$3.14* & {85.26}$\pm$4.12* & \textbf{94.57}$\pm$1.39* & {83.51}$\pm$4.37 & {91.16}$\pm$3.41* &88.02 \\
CCE + ProxyNCA    &88.19$\pm$1.91*&85.60$\pm$2.79*&83.84$\pm$4.96&94.01$\pm$1.23&81.16$\pm$4.91&90.22$\pm$3.22&87.17\\
CCE + ProxyAnchorLoss   &87.26$\pm$2.31 &84.18$\pm$3.74 &85.63$\pm$4.32*&93.81$\pm$1.62&81.60$\pm$4.94&89.18$\pm$3.51&86.94\\
CCE + SoftTriple   + inf   &88.24$\pm$2.22* &\textbf{85.66}$\pm$2.95*&83.51$\pm$5.81&94.22$\pm$1.73&84.26$\pm$4.22&90.71$\pm$3.60&87.76\\
CCE + ProxyNCA  + inf   &88.33$\pm$1.67* &85.52$\pm$2.89* &86.14$\pm$4.23*&93.95$\pm$1.48&84.08$\pm$4.40&90.01$\pm$3.93&88.01\\
CCE + ProxyAnchorLoss   + inf   &\textbf{88.68}$\pm$1.39* &\textbf{85.66}$\pm$2.95* &\textbf{87.95}$\pm$3.23*&94.10$\pm$1.63&\textbf{84.65}$\pm$4.35*&\textbf{91.71}$\pm$2.04*&\textbf{88.79}\\

\end{tabular}
}
\label{tab:rl-100-examples-dataset}
\end{table*}

\begin{table*}[h!]
\centering
\caption{F1 score of RoBERTa-large (RL) vs RoBERTa-large with DML losses trained on the 1,000-element datasets.}
\resizebox{\columnwidth}{!}{%
\begin{tabular}{llllllll}
\hline
\textbf{Loss} &                      \textbf{SST2}                  & \textbf{MR}                  & \textbf{MPQA}         & \textbf{SUBJ}         & \textbf{TREC}         & \textbf{CR}           &  \textbf{Avg}\\
\hline
CCE               & 91.59$\pm$0.69          & 89.73$\pm$2.20       & 90.16$\pm$1.25  & 96.04$\pm$1.30 & 89.89$\pm$2.87 & 93.16$\pm$2.60  & 91.76\\
CCE + Triplet   &91.62$\pm$0.71&89.56$\pm$2.33 &90.25$\pm$2.33&95.94$\pm$1.41&88.89$\pm$2.74&93.17$\pm$3.14&91.57\\
CCE + NPairs   &91.81$\pm$0.85 &89.59$\pm$2.20&87.35$\pm$7.35*&96.22$\pm$1.16&88.53$\pm$2.92*&93.56$\pm$2.48&91.18\\
CCE + SupCon      & {91.98}$\pm$0.70*  &89.91$\pm$2.17      & 89.94$\pm$1.95     & 96.23$\pm$1.16  & {90.34}$\pm$2.66     & 93.90$\pm$2.63    &  92.05\\
CCE + SoftTriple  &{91.98}$\pm$0.81*&{90.02}$\pm$2.00&{90.60}$\pm$1.74& {96.26}$\pm$1.20 & 89.93$\pm$2.45    & {94.00}$\pm$2.28 &92.13\\
CCE + ProxyNCA   &92.32$\pm$0.62*&\textbf{90.33}$\pm$2.10&91.01$\pm$1.62*&96.32$\pm$1.11&90.92$\pm$1.90&93.85$\pm$2.34 & 92.45\\
CCE + ProxyAnchorLoss  &92.37$\pm$0.65* &90.22$\pm$2.19&90.86$\pm$1.77*&96.29$\pm$1.06&91.16$\pm$2.46*&93.75$\pm$2.86 & 92.44\\
CCE + SoftTriple   + inf   &92.12$\pm$0.69*&90.28$\pm$2.29&\textbf{91.04}$\pm$1.60* &96.29$\pm$1.28&92.62$\pm$2.13*&94.02$\pm$2.90 & 92.73\\
CCE + ProxyNCA  + inf    &92.35$\pm$0.79* &90.23$\pm$2.18&90.81$\pm$1.80&96.28$\pm$1.37&92.85$\pm$2.23*&93.85$\pm$2.67 & 92.73\\
CCE + ProxyAnchorLoss   + inf    &\textbf{92.54}$\pm$0.68*&90.22$\pm$2.31&90.97$\pm$1.34*&\textbf{96.42}$\pm$1.20&\textbf{93.23}$\pm$2.05*&\textbf{94.27}$\pm$2.33* & \textbf{92.94}\\
\end{tabular}
}
\label{tab:rl-1000-examples-dataset}
\end{table*}



\section{Experimental Results}
\label{sec:experiments}

\Cref{tab:rb-20-examples-dataset,tab:rb-100-examples-dataset,tab:rb-1000-examples-dataset,tab:rl-20-examples-dataset,tab:rl-100-examples-dataset,tab:rl-1000-examples-dataset} present our results separately for the RoBERTa-base and RoBERTa-large encoders trained on $20$, $100$ and $1,000$ observations. In the results, we included the baseline performance, where the models were trained with \textit{CCE} loss as well as models trained with the different \textit{DML} losses. The results also include the performance of the models with the modified inference process denoted with \textit{+inf} in tables, as described in Equation~\ref{eq:novel-logits}. We also calculated p-values for the results, and placed a $*$ sign next to the results for which the p-value was less than 0.05.

\subsection{Non-proxy-based DML Loss}\label{subs:non-proxy-based-dml}

Models trained with the non-proxy DML losses, i.e. \textit{Triplet Loss} \textit{N-Pairs Loss} and \textit{SupCon Loss}, on average outperform the baseline by 0.97 percentage points but are worst than the proxy-based DML losses and proxy-based DML with inference modification by 0.98 and 1.57 percentage points, respectively. The best non-proxy DML loss is the \textit{SupCon Loss}. It is better on average than the baseline by 1.58 percentage points. The \textit{N-Pair Loss} is better on average than the baseline by  0.42 percentage points and the \textit{Triplet Loss} by 0.90 percentage points. As we can see, surprisingly, the worst non-proxy-based DML loss and worst of all analysed DML losses is the \textit{N-Pairs Loss}. Moreover, considering the average F1 score for RoBERTa-base and 100-element datasets, the \textit{SupCon Loss} obtained the highest score among all losses, is marginally better than the \textit{ProxyAnchor + inf}. Also, we note that the p-value is greater than 0.05 for most of the results in the non-proxy group.

\subsection{Proxy-based DML Loss}
Models trained with proxy-based DML losses, i.e. \textit{SoftTriple Loss} \textit{ProxyNCA Loss} and \textit{ProxyAnchor Loss}, are on average better than the baseline by 1.94. The \textit{SoftTriple Loss} had the highest average performance increase from baseline among all proxy-based DML losses, at 2.24 percentage points. The \textit{ProxyAnchor Loss} increased performance by 1.91 percentage points, and \textit{ProxyNCA Loss} noted the  performance gain at 1.68 percentage points. Furthermore, the p-value is less than 0.05 for most of the losses in this group when the dataset size is 20 or 100. Furthermore, if we consider only \textit{SofTriple Loss}, 63\% of the results from data sets with 20 and 100 observations have a p-value below 0.05.

\subsection{Proxy-based DML Loss with Inference Modification}
The language models fine-tuned with the proxy-based DML losses and inference modification yielded the best performance gain, averaging 2.54 percentage points over the baseline model, of all DML losses tested.

Modification of the inference yielded further performance gains, with models outperforming the baseline by 3.27, 2.00, and 2.35 percentage points for \textit{ProxyAnchor Loss + inf}, \textit{ProxyNCA Loss + inf}, and \textit{SoftTriple Loss + inf}, respectively. That is, the inference modification increased the performance of all proxy-based DML losses, with the highest gain reported for \textit{ProxyAnchor Loss} at 1.36 percentage points, followed by 0.33 percentage point for \textit{ProxyNCA Loss} and 0.11 percentage points for \textit{SoftTriple Loss}.

Based on the results, we found that, on average, the best DML loss was \textit{ProxyAnchor + inf}, which increased the performance of the fine-tuned models by 3.27 percentage points. The second loss was \textit{SoftTriple + inf} resulting in the performance gain of 2.35 percentage points, and the third \textit{SoftTriple} increased the performance by 2.24 percentage points.

As in the previous case, the p-value is less than 0.05 in most cases when the dataset size is 20 or 100. However, when considering only \textit{ProxyAnchor + inf}, about 64\% of all the results have p-value smaller than 0.05. If we consider only datasets of size 20 and 100, 75\% of the results have p-value smaller than 0.05.

\subsection{Few-shot Learning Settings}
\begin{figure}[!ht]
    \centering
    \includegraphics[scale=0.6]{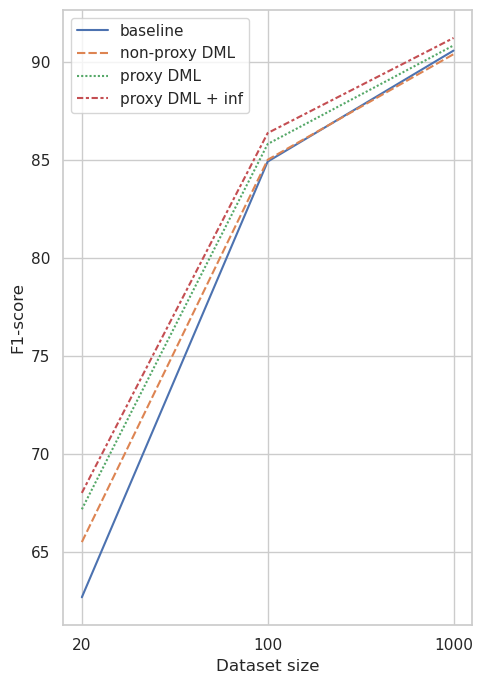}
    \caption{Performance comparison between different families of DML methods and the baseline.}\label{fig:performance_comparison}
\end{figure}

The previous analysis shows that the models fine-tuned with DML losses on average perform better than the baseline models. Still, this increase is not uniform across different dataset sizes. The average performance increase over the baseline model for datasets of size 1,000 for all DML losses is about 0.22 percentage points, with the most significant increase for the \textit{ProxyAnchor + inf} at about 0.78 percentage points. For a 100-element dataset, the average gain is 0.73 percentage points, with the largest performance increase for \textit{ProxyAnchor + inf} at 1.89 percentage points. For the 20-element dataset, there was an average increase over baseline of about 3.95 percentage points, with the most significant growth for \textit{ProxyAnchor + inf} at about 7.14 percentage points. The performance overview of different DML family methods throughout dataset size is shown in Figure~\ref{fig:performance_comparison}.

Our results show that the smaller the dataset, the more performance gains can be obtained using DML losses. We also note that the highest performance increase can be gained using the proxy-based DML loss function with the inference modification.

\section{Conclusion}\label{sec:conclusion}

In this paper, we investigated the influence of DML loss functions' performance during supervised fine-tuning of RoBERTa-base and RoBERTa-large language models compared with baselines (fine-tuned with the use of CCE loss only). We also studied the impact of modifying the inference procedure on the models' performance. The analysis was performed based on the 40-fold cross-validation over several datasets from the SentEval Transfer Tasks in the few-shot learning settings. We found that each DML loss function, on average, increases the performance of the RoBERTa base and large encoders. The non-proxy-based DML losses improve on average the performance by 0.97 percentage points, with the highest increase for \textit{SupCon Loss} at 1.58 percentage points. The proxy-based DML losses increase the model's performance by 1.94 percentage points, with the highest performance gain for the \textit{SoftTriple Loss} at 2.24 percentage points. In addition, applying inference modifications to models fine-tuned with proxy-based DML losses steadily improves the performance by an average of 0.6 percentage points, with the most significant gain for \textit{ProxyAnchor + inf} being an increase of 3.27 percentage points over the baseline, making this loss the best of all tested.

\bibliographystyle{unsrtnat}
\bibliography{references}  






\end{document}